\documentclass[conference]{IEEEtran}
\IEEEoverridecommandlockouts
\usepackage{lineno}
\usepackage{cite}
\usepackage{amsmath,amssymb,amsfonts}
\usepackage{algorithmic}
\usepackage{graphicx}
\usepackage{textcomp}
\usepackage{xcolor}
\usepackage{cleveref}
\usepackage{booktabs}
\usepackage{url}
\usepackage[square,numbers]{natbib}
\usepackage{caption}
\makeatletter
\renewcommand{\paragraph}{%
  \@startsection{paragraph}{4}%
  {\z@}{0.2em}{-1em}%
  {\normalfont\normalsize\bfseries}%
}
\makeatother

\newcommand{\diag}{\operatorname{diag}}
\newcommand{\rank}{\operatorname{rank}}
\newcommand{\agg}{\operatorname{agg}}

\usepackage{lipsum}  
\usepackage{mathtools}

\usepackage{colortbl}
\usepackage{xcolor}
\usepackage{pgfplots}

\newcommand{\applygradient}[3]{%
    \pgfmathsetmacro{\percent}{100.0*(#1-#2)/(#3-#2)}%
    \edef\tempcolor{\noexpand\cellcolor{red!\percent!white}}%
    \tempcolor
}

\begin{document}

\title{\bfseries \LARGE Training-free Neural Architecture Search \\
through Variance of Knowledge of Deep Network Weights}

\author{Ondřej Týbl\\
Department of Cybernetics\\
FEE, Czech Technical University\\
{\tt\small tyblondr@cvut.cz}
\and
Lukáš Neumann\\
Department of Cybernetics\\
FEE, Czech Technical University\\
{\tt\small lukas.neumann@cvut.cz}
}

\maketitle

\begin{abstract}
Deep learning has revolutionized computer vision, but it achieved its tremendous success using deep network architectures which are mostly hand-crafted and therefore likely suboptimal. Neural Architecture Search (NAS) aims to bridge this gap by following a well-defined optimization paradigm which systematically looks for the best architecture, given objective criterion such as maximal classification accuracy. The main limitation of NAS is however its astronomical computational cost, as it typically requires training each candidate network architecture from scratch.

In this paper, we aim to alleviate this limitation by proposing a novel training-free proxy for image classification accuracy based on Fisher Information. The proposed proxy has a strong theoretical background in statistics and it allows estimating expected image classification accuracy of a given deep network without training the network, thus significantly reducing computational cost of standard NAS algorithms. 

Our training-free proxy achieves state-of-the-art results on three public datasets and in two search spaces, both when evaluated using previously proposed metrics, as well as using a new metric that we propose which we demonstrate is more informative for practical NAS applications. The source code is publicly available at {\small\url{https://www.github.com/ondratybl/VKDNW}}.

\end{abstract}
\vspace{-5pt}
\section{Introduction}
\label{sec:intro}

In most instances, neural network architectures are designed by authors following the field's ``best-practices'' or their experience, without any formal and repeatable procedure. This is however inconvenient especially in applications on a large scale. Neural Architecture Search (NAS) aims to bridge this gap by following a well-defined optimization paradigm which systematically looks for the best architecture, given objective criterion such as maximal accuracy.

\begin{figure}
    \centering    
    \includegraphics[width=0.95\columnwidth]{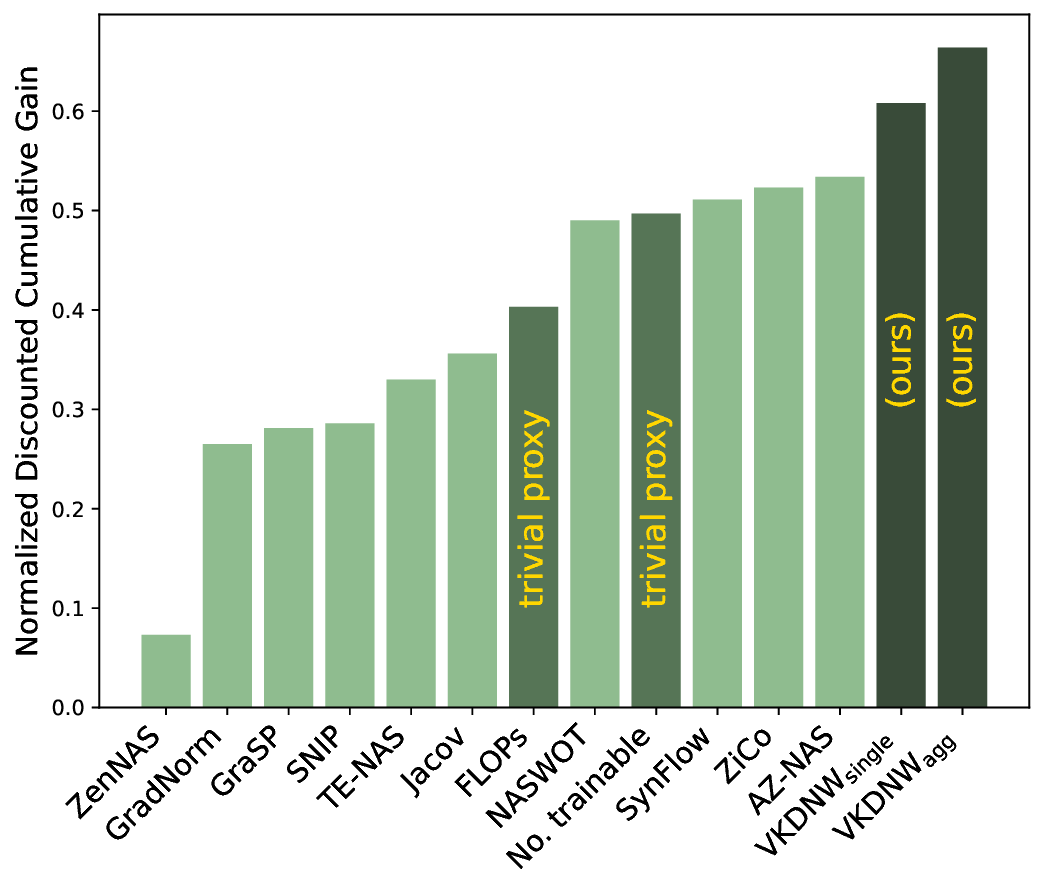} \vspace{-5pt}
    \caption{Training-free NAS methods on 
ImageNet16-120 \cite{dong2020bench}. Methods are compared by Normalized Discounted Cumulative Gain (see Sec. \ref{sec:evaluationmetrics}), our method (VKDNW) is the best also measured by Kendall's $\tau$ and Spearman's $\rho$ correlations (see Table \ref{tab:table1}). Also note that simple \textit{number of trainable layers} (below denoted $\aleph$) is significantly better trivial proxy than the number of FLOPs.
\vspace{-5pt}}
    \label{fig:example_bad_kendall}
\end{figure}

The main limitation of Neural Architecture Search (NAS) is however the computational cost, because in the most basic NAS setup, it is required to train thousands or more of different deep network architectures from scratch in order just to calculate a single scalar  -- the objective function value, such as the classification accuracy. This severely limits practical applications of NAS as the size of feasible architecture search space is only a small fraction of the overall space of all networks.

Training-free Neural Architecture Search (TF-NAS) aims to alleviate this limitation by introducing an \textit{objective function proxy} which -- unlike the actual objective function -- does not require training the network. As a result, a good proxy allows the TF-NAS algorithm to explore significantly bigger portion of the network architecture search space compared to traditional NAS, and to find the best network architecture without training a single network. The crucial question is however finding an appropriate objective function proxy.

In this paper, we present a novel, principled \textit{objective function proxy} called \textit{Variance of Knowledge of Deep Network Weights (VKDNW)} for image classification accuracy, which allows us to \textbf{find optimal network architectures for image classification without training} them first. Our method is, to the best of our knowledge, the first successful application of Fisher Information theory~\cite{ly2017tutorial} in the context of large deep neural networks, and as such allows us to formally describe and quantify the difficulty of network parameters' estimation process. In other words, \textit{given a network architecture, our method estimates how easy or hard it will be to train the network}.

Additionally, we also observe that the evaluation metrics used in the TF-NAS community are not well-suited to the problem at hand, because it unnecessarily penalizes for bad proxy accuracy for networks which are not interesting, and vice-versa it does not sufficiently reward proxies which are able to accurately pick out good network architectures. Following this observation, we propose that the Normalized Discounted Cumulative Gain should be used in companion with other TF-NAS metrics, and show indeed that there are significant differences amongst previously proposed TF-NAS methods when the new metric is considered. 

To summarize, we make the following contributions:
\begin{enumerate}
    \item We introduce a novel algorithm for estimation of Fisher Information Matrix spectrum, which is tractable even for models with large number of parameters such as deep networks, that overcomes the usual problems of numerical stability.
    \item We introduce a novel principled VKDNW proxy for image classification accuracy. The proxy is based on strong theoretical background and captures uncertainty in weight estimation process. It brings information that is orthogonal to the model size which then allows for efficient combination with previously proposed proxies, leading to state-of-the-art results.
    \item We propose a new evaluation metric for TF-NAS proxies which is more relevant to the actual NAS objective as it concentrates on ability of given proxy to identify good networks.
\end{enumerate}


\section{Method}
\label{sec:method}

Our zero-shot proxy for image classification accuracy builds on Fisher Information theory~\cite{ly2017tutorial}, therefore we begin by a thorough analysis of Fisher Information Matrix (FIM) of the network weights estimation problem (see Sec. \ref{subsec:fim}). We proceed by discussing challenges involved in practical application of FIM in context of large over-parametrized models that lead to only limited success in previous works and present our contributions to overcome these limitations (see Sec.~\ref{subsec:empiricalFIM}). Finally. we propose a novel FIM-based proxy for NAS algorithms (Sec. \ref{subsec:vkdnw}).

\subsection{Fisher Information}
\label{subsec:fim}

The problem of finding the optimal weights of a neural network $f$ for the task of $C$-class image classification can be seen as a maximum likelihood estimation with a statistical model
\begin{align}
    \sigma_\theta(c\,|\,x)=\frac{\exp\left({\Psi_c(x, \theta)}\right)}{\sum_{d=1}^C\exp\left({\Psi_d(x, \theta)}\right)}, \quad c=1,\dots, C
\end{align}
where $\sigma_\theta(\cdot\,|\,x)$ denotes the a posteriori distribution of the labels given input image $x$ and $\Psi(x,\theta)\in\mathbb{R}^C=\left( \Psi_1(x,\theta),\dots ,\Psi_C(x,\theta)\right)$ is the network output (logits) given the weight vector $\theta\in\mathbb{R}^p$, i.e. the network weights. 
We describe the process of training as finding the optimal weight vector $\theta^*$ that fits our data and \textit{we posit that network architectures should be characterised by how easy it is to estimate their optimal network weights} $\theta^*$. 
We build upon statistical learning theory and use Fisher Information~\cite{ly2017tutorial} framework to formally describe expected behaviour of the training process of a given deep network $f$.

The Fisher Information Matrix (FIM) encompasses information on the difficulty of the parameter estimation problem and it plays a crucial role in several fundamental results, which we apply below in the context of deep networks. The FIM of a network $f$ and its set of weights  $\theta\in\mathbb{R}^p$ is given as
\begin{align}
\label{eq:fim}
F(\theta)\coloneq\mathbb{E}\left[\nabla_{\theta}\sigma_\theta(c\,|\,x)\,\nabla_{\theta}\sigma_\theta(c\,|\,x)^T\right]\in\mathbb{R}^{p\times p},
\end{align}
where we take the expected value $\mathbb{E}$ with respect to the joint distribution of $(x,c)$. For more detailed account on Fisher Information theory and its applications in the context of  machine learning, we kindly refer reader to \cite{karakida2019universal, martens2020new, lee2022masking, pennington2018spectrum, park2019adaptive}.

\paragraph{Cramér-Rao bound.} The first part of estimation theory we build upon is the inverse of the FIM, known as the Cramér–Rao bound (see \cite{frieden2010exploratory}), which is the asymptotic variance of the estimated weights (i.e. the uncertainty coming from the data variability). Thus, the larger the matrix norm of the FIM, the more certain we are about the weights $\theta$. More specifically, any data-dependent estimator $\Hat{\theta}_n=\Hat{\theta}_n(x_1,\dots,x_n)$ is a random vector with randomness coming from the (independent) choice of input images $x_1,\dots,x_n$ and as such has some variance matrix $\text{Var}\left(\Hat{\theta}_n\right)$. The famous result (see \cite{cramer1999mathematical, rao1992information}) named in honour of H. Cramér and C. R. Rao states that if $\Hat{\theta}_n$ is unbiased then the variance is bounded from below as
\begin{align}
    \text{Var}\left(\Hat{\theta}_n\right)\geq \frac{1}{n}F^{-1}(\theta)
\end{align}
and for maximum likelihood estimation this bound is attained as the number of input images grows to infinity $n\to\infty$. 

We formally show  (see Supplementary material) that  for each weight $\theta(j)$ the mean square error of our estimation is controlled by the diagonal element of the FIM inverse as
\begin{align}
    \mathbb{E}\left( \Hat{\theta}_n(j)-\theta(j)\right)^2\geq \frac{1}{n}\left(F^{-1}(\theta)\right)_{jj}.
\end{align}
Therefore, knowing the FIM allows us to evaluate how certain we are about the weight estimates. We can go even further by inspecting the eigenvalues of $F(\theta)$. Denoting the largest and smallest eigenvalue of the FIM as $\lambda_{\min}$ and $\lambda_{\max}$ respectively, then we have that there exist linear combination coefficients $e_{\min}$ and $e_{\max}$ of unit size so that
\begin{align}
    \mathbb{E}\left( e_{\min}^T\Hat{\theta}_n-e_{\min}^T\theta\right)^2\geq &\frac{1}{n\lambda_{\min}},\nonumber\\ 
    \mathbb{E}\left( e_{\max}^T\Hat{\theta}_n-e_{\max}^T\theta\right)^2\geq &\frac{1}{n\lambda_{\max}},
\end{align}
indicating that if the difference between the largest and smallest eigenvalue is \textit{large} then there exist combinations of weights with \textit{very large difference} in the estimation certainty. Altogether, the more the eigenvalues of the FIM are similar, the more similar is also the variance in the weight estimation across all model weights.

\paragraph{Prediction sensibility.}
The change of the model prediction measured by the KL-divergence when subject to a small perturbations of the weights is given as
\begin{align}
    D_{KL}(\sigma_{\theta+\theta_\delta}(\cdot\,|\,x), \sigma_{\theta}(\,\cdot|\,x))\approx\frac{1}{2}\theta_\delta^T\,F(\theta)\,\theta_\delta
\end{align}
for a small weight perturbation vector $\theta_\delta\in\mathbb{R}^p$. Thus, $F(\theta)$ measures volatility of the predictions subject to the weight change -- if the difference between the largest and the smallest eigenvalues of the FIM is large, then there exist perturbation directions $\theta_{min}, \theta_{max}$ corresponding to $\lambda_{min}, \lambda_{max}$ such that
\begin{align}
    D_{KL}&(\sigma_{\theta+\theta_{min}}(\cdot\,|\,x), \sigma_{\theta}(\cdot\,|\,x)) 
    \approx\frac{1}{2}\theta_{min}^T\,F(\theta)\,\theta_{min} = \nonumber\\
    &=\frac{1}{2}\lambda_{min}\|\theta_{min}\|^2\ll\frac{1}{2}\lambda_{max}\|\theta_{max}\|^2 = \\
    &=\frac{1}{2}\theta_{max}^T\,F(\theta)\,\theta_{max}
    \approx D_{KL}(\sigma_{\theta+\theta_{max}}(\cdot\,|\,x), \sigma_{\theta}(\cdot\,|\,x)) \nonumber
\end{align}
and therefore in some directions a small change of the weights has much larger impact on the prediction than in others, making the model less balanced. For further discussion, see the Supplementary material.

\subsection{Empirical Fisher Information Matrix implementation}
\label{subsec:empiricalFIM}
When independent and identically distributed sample images $x_n$ are available, empirical FIM $\Hat{F}(\theta)$ is defined as
\begin{align}
\Hat{F}(\theta)&\coloneq\frac{1}{n}\sum_{n=1}^N\mathbb{E}_{\sigma_\theta}\left[\nabla_{\theta}\sigma_\theta(c\,|\,x_n)\,\nabla_{\theta}\sigma_\theta(c|\,x_n)^T\right]
    \label{eq:fim_epirical}
\end{align}
where $\mathbb{E}_{\sigma_\theta}$ now denotes the expectation with respect to the model prediction $\sigma_\theta$.

We would like to first emphasize several crucial aspects and \textit{contributions of this paper} that lead to the first success of the Fisher Information theory (having e.g. \cite{abdelfattah2021zero} in mind) in the context of Neural Architecture Search, despite Fisher Information being one of the first obvious choices for deep network analysis:

1. Following \cite{kunstner2019limitations} we write the empirical FIM \eqref{eq:fim_epirical} as
\begin{align}
    \Hat{F}(\theta)=\frac{1}{n}\sum_{n=1}^{N}&\left[\nabla_\theta\Psi (x_n,\theta)^T\left(\diag(\sigma_\theta(\cdot, x_n))-\right.\right.\nonumber\\
    &\left.\left.\sigma_\theta(\cdot, x_n)\sigma_\theta(\cdot, x_n)^T\right)\nabla_\theta\Psi (x_n,\theta)\right].
\end{align}
and we further decompose the inner matrix $\diag(\sigma_\theta(\cdot, x_n))-\sigma_\theta(\cdot, x_n)\sigma_\theta(\cdot, x_n)^T$ using analytical formulas from \cite{tanabe1992exact} to avoid numerical instability of the computation as we arrive at a feasible representation
\begin{align}
    \Hat{F}(\theta)=\frac{1}{n}\sum_{n=1}^{N}A_n^TA_n
\end{align}
for some matrices $A_n\in\mathbb{R}^{C\times p}$. We observed that networks typically yield very imbalanced outputs at initialization (i.e. every network prioritizes few classes over the rest) making it extremely important not to exclude the factor $\diag(\sigma_\theta(\cdot, x_n))-\sigma_\theta(\cdot, x_n)\sigma_\theta(\cdot, x_n)^T$, which is however difficult to compute without underflow/overflow.

2. The dimension of FIM is equal to $p$ (the number of all trainable parameters) and therefore direct computation of the eigenvalues is numerically intractable and unstable. However, our results show that if a small number of representative parameters is drawn, then computation becomes stable while the discrimination power does not suffer. We used a simple rule where a single parameter from each trainable layer (not including batch normalization) is chosen. Stability with respect to choice of such parameters can be found in Supplementary material.
    
3. FIM of large networks typically suffers from having pathological spectrum, i.e. there usually exist zero eigenvalue and its multiplicity is large (see \cite{karakida2019pathological}) and thus the eigenvalues estimation due to a large condition number is imprecise. However, as we deal with a symmetric positive-semidefinite matrix, the eigenvalues actually coincide with the singular values (see \cite{leon2006linear}), for which the estimation algorithm performs better.

Let us also emphasize that there is a common misconception in part of the community as it is often mistakenly assumed that one can simply use the true labels of $x_n$ in \eqref{eq:fim_epirical} in place of $c$. However, such definition is then meaningless as it does not approximate the FIM in the classical Monte Carlo sense (see Supplementary material and \cite{kunstner2019limitations} for a comparison). That means that the empirical FIM \textit{does not} depend on the true labels and therefore our method \textit{does not} require real data - we use random input instead.

\subsection{Variance of Knowledge for Deep Network Weights}
\label{subsec:vkdnw}

In order to characterize properties of the parameter estimation process for a given network $f$ through the lens of Fisher Information theory, we inspect the eigenvalues of the empirical FIM $\Hat{F}(\theta_{\text{init}})$ and define the entropy of \textit{Variance of Knowledge for Deep Network Weights (VKDNW)} as
\begin{align}
\label{eq:vkdnw}
    \text{VKDNW}\big(f\big) &\coloneq -\sum_{k=1}^9\tilde{\lambda}_k\log\Tilde{\lambda}_k \nonumber\\
    \tilde{\lambda}_k&=\frac{\lambda_k}{\sum_{j=1}^9\lambda_j}, \quad k=1,\dots, 9,
\end{align}
where $\lambda_k$ denotes the $k$-th decile of the FIM eigenvalues as the representation of the FIM spectrum\footnote{As explained in Sec. \ref{subsec:empiricalFIM}, the smallest eigenvalue $\lambda_0$ is usually equal to zero and thus we exclude it for stability reasons (similarly with the maximal eigenvalue $\lambda_{10}$)}, and $\theta_{\text{init}}$ denotes network weights at initialization.

Our score therefore measures the diversity of the FIM eigenvalues, and from the entropy theory we know that VKDNW attains its maximum exactly when all the eigenvalues $\lambda_k$ are equal and VKDNW gets lower as the eigenvalues become \textit{more different}. Based on the discussion in Sec. \ref{subsec:fim} we see that VKDNW is high when the uncertainty in all model weight combinations are similar (see Cramér-Rao bound) and there are no directions in the weight space that would influence the network prediction substantially differently than others. Due to the fact that we have normalized both the number of eigenvalues under consideration (by taking a fixed number of representatives irrespective of the network size) and the magnitude of the eigenvalues, VKDNW is independent of network size (number of network weights $p$).

Let us note, that we are familiar with the fact that even though our motivation was (among others) based on Cramér-Rao bound that assumes evaluation of FIM at the \textit{correct} weight vector $\theta_X$ that fits the data, which is typically far away from the weights given at initialization. However, our empirical results below support the hypothesis that the evaluation despite being in the wrong point brings valuable information.

\paragraph{Ranking networks for NAS.}
The proposed VKDNW score is independent of network size, which is extremely beneficial to compare individual structures of network architectures. Thus, it does not aim to capture capacity, rather it targets feasibility of the computation graph given number of operations. However, when comparing different network structures and different network sizes together as it is done in NAS, one indeed needs to take network size into account as well, because naturally larger networks have bigger capacity and therefore tend to have higher accuracy.  To capture also the capacity we proxy the network size by the number of layers with weights that we denote as $\aleph \big(f\big)$ for a network $f$ and we introduce the ranking
\begin{align}
\text{VKDNW}_{\text{single}}\big(f\big)\coloneq\aleph\big(f\big)+\text{VKDNW}\big(f\big)
\label{eq:vkdnwrank}
\end{align}
%
Here we leverage the fact that VKDNW as an entropy of some quantity is always between 0 and 1 and by summing it with an integer-valued quantity we in fact obtain that we have first grouped networks by our size proxy $\aleph$ and then within each group of similar networks sizes we order them by VKDNW. 

\section{Evaluation Metrics}
\label{sec:evaluationmetrics}

\begin{figure}
    \centering
    \includegraphics[width=\columnwidth]{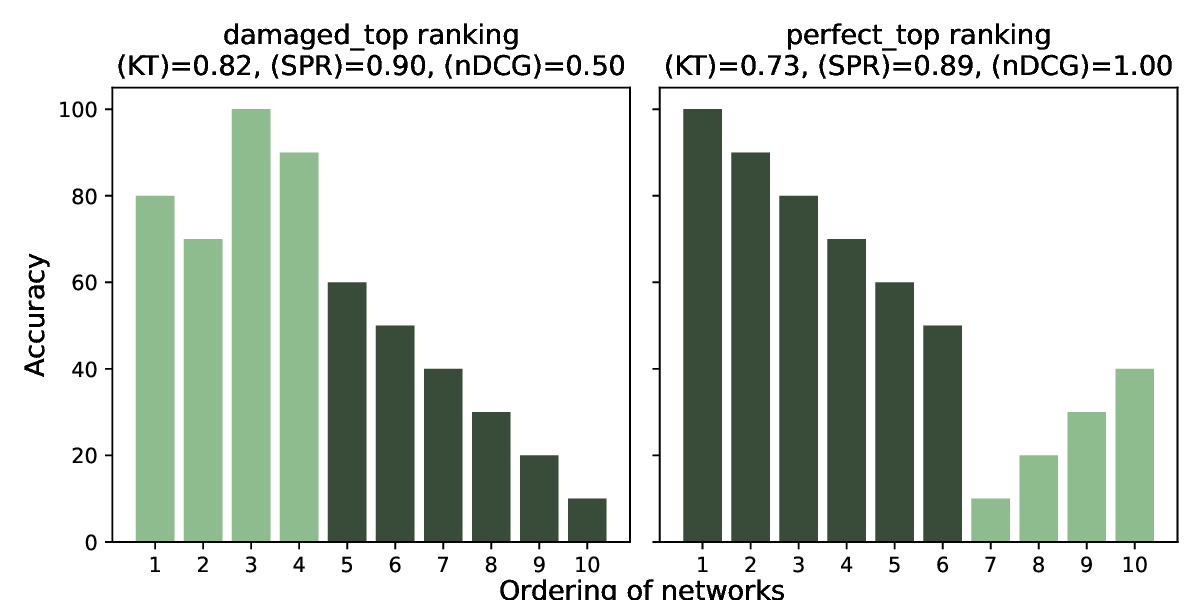}
    \caption{Toy example of two rankings on 10 networks. We plot accuracies ordered by the rankings and evaluation metrics Kendall's $\tau$ (KT) and Spearman's $\rho$ (SPR) correlations and Normalized Discounted Cumulative Gain ($\text{nDCG}_{5}$). \vspace{-15pt}}
    \label{fig:toy_example_raking}
\end{figure}
In this section, we dive into the problem of evaluating (image) classification accuracy proxies for training-free NAS (TF-NAS). 
In NAS algorithms, the proxy actually does not have to predict the classification accuracy in absolute terms -- since we are after finding the ``best'' network for given task -- it's sufficient that the proxy properly ranks the networks, ordering them from worst to the best.

Suppose that a collection of $K$ networks with validation accuracies $\text{acc}_1, \dots, \text{acc}_K$ have been ranked by a TF-NAS proxy as $r_1,\dots,r_K\in\mathbb{R}$. The standard way of evaluating~\cite{li2023zico, lee2024az,kadlecova2024surprisingly} how well the ranks correspond to the accuracies is to compute Kendall’s $\tau$ (KT, \cite{kendall1938new}) and Spearman’s $\rho$ (SPR, \cite{spearman1961proof}) rank correlation coefficients. Kendall's $\tau$ is given as
\begin{align}
    \tau\coloneq\frac{n_c-n_d}{\sqrt{n_c+n_d+n_1}\sqrt{n_c+n_d+n_2}},
\end{align}
where $n_c$ denotes the number of concordant pairs $\left(\text{acc}_k, r_k\right)$, $n_d$ the number of discordant pairs and $n_1$ (resp. $n_2$) denote the number of ties in $\text{acc}_k$ only (resp. in $r_k$ only). In our context, (KT) relates after some normalization to a probability that two randomly chosen network rankings $r_k, r_l$ are ordered correctly according to their accuracies $\text{acc}_k, \text{acc}_l$. In contrast, (SPR) is defined to account more for outliers, i.e. it tends to put a large penalization if there are some networks for which the difference between the rank of the accuracy $\text{acc}_k$ and the rank $r_k$ is large as it is just the classical correlation coefficient however applied on the orders of the assessed quantities.

\paragraph{Normalized Discounted Cumulative Gain.}
Both (KT) and (SPR) are good evaluation metrics when our interest lies in comparison of ranking proxies when \textit{all networks involved are of the same importance}. However, in NAS we are primarily interested if a ranking helps us to pick best networks from a given collection and less interested how well the ranking compares two networks with low accuracy. 
Finding inspiration in information retrieval where one measures the quality of a system that retrieves resources relevant to an input query, we propose to use Normalized Discounted Cumulative Gain (nDCG)~\cite{burges2005learning} as a more relevant metric to measure quality of TF-NAS proxies. The nDCG metric is defined with a key requirement that highly relevant documents are more valuable when they appear earlier in the search engine results (i.e., in higher-ranking positions). This requirement can be reformulated for the task of neural architecture search as \textit{networks with high accuracies are more valuable when they appear on higher-ranking positions}. 

We thus define the metric as follows: first we order the networks by their ranking $r_1,\dots, r_K$ so that $r_{k_1}=K, r_{k_2}=K-1,\dots$ and we compute
\begin{align}
    \text{nDCG}_{P}\coloneq\frac{1}{Z}\left(\sum_{j=1}^P\frac{2^{\text{acc}_{k_j}}-1}{\log_2\left(1+j\right)}\right),
    \label{eq:gain}
\end{align}
where $P\in\mathbb{N}$ is a parameter determining how many top-ranked networks do we consider (e.g. it corresponds to the population size in an evolution algorithm) and $Z$ is a normalization factor that represents the ideal discounted cumulative gain so that nDCG is equal to one for a perfect fit. 

The higher nDCG the better the ranking is as it is a weighted average of transformed top-ranked accuracies. In \cite{yining2013theoretical} it is shown that the choice of the discount factor of given by inverse logarithm is a good choice as it nDCG then well separates different ranking systems.
\footnote{In case of ties we take random ordering within the groups (parameter \texttt{ignore\_ties=True} in the scikit-learn implementation).}

\paragraph{Toy Example.} Let us briefly demonstrate the weak ability of (KT) and (SPR) to distinguish rankings that poorly discriminate networks with high accuracy. Suppose a collection of $10$ networks with their validation accuracies 
$100, 90, \cdots, 10$ is given and our aim is to rank them from worst to the best. We evaluate two different toy rankings: a) ranking $\texttt{damaged\_top}$ which perfectly fits the accuracies, only the two top networks are swapped with the third and fourth, b) ranking $\texttt{perfect\_top}$ which fits perfectly for the top-performing networks, however, now the order of the four worst networks is reversed (see Figure~\ref{fig:toy_example_raking}). The $\texttt{damaged\_top}$ ranking is an example that should be evaluated worse for the architecture search task compared to $\texttt{perfect\_top}$ as in the first case the best networks are not placed first, while in the second they are. On the other hand, even though $\texttt{perfect\_top}$ is not a perfect fit, it still discriminates the top networks ideally and o\textit{nly struggles for networks with low accuracy}, which are not interesting for NAS which is after the best networks. From the correlation perspectives of (KT), (SPR) $\texttt{damaged\_top}$ is better than $\texttt{perfect\_top}$, therefore, if we rely only on these two evaluation metrics we prefer ranking that is worse for the architecture search task as the top networks are ranked worse. On the other hand, for $\texttt{perfect\_top}$ we obtain ideal $\text{nDCG}_{5}$ while it drops to 0.5 for $\texttt{damaged\_top}$. We conclude that using $\text{nDCG}_{5}$ we choose ranking that has higher discriminatory power for good networks. 

We also note that from statistical perspective the (KT) and (SPR) of $\texttt{damaged\_top}$ is significantly higher than of a random ranking\footnote{That is, the $p$-value for the hypothesis that $\texttt{damaged\_top}$ is assigned independently of the accuracies is below 0.005.}, and therefore we'd conclude that $\texttt{damaged\_top}$ is not independent of ground truth accuracy. On the other hand, when we perform the same statistical test on $\texttt{damaged\_top}$ using $\text{nDCG}_{5}$, we conclude that the $\texttt{damaged\_top}$ is not significantly better than ranking networks randomly\footnote{Running 1000 samples on a random ranking we obtain that $\texttt{damaged\_top}$ has $\text{nDCG}_{5}$ around 75th percentile of such random evaluations and it is not significantly better than a random ranking.}.

\section{Related Work}
\label{sec:relatedwork}
\vspace{-5pt}

Zero-shot NAS aims to rank given networks in a training-free manner based on their (a priori unknown) final performance, which allows to prune the huge search space with limited costs when seeking for optimal architecture \cite{ying2019bench, dong2020bench, liu2018darts} and other configuration~\cite{cai2018proxylessnas, lin2021zen, sandler2018mobilenetv2}.

\begin{table*}[t]
    \centering    
\begin{tabular}{lc|ccc|ccc|ccc}
\hline
 & & \multicolumn{3}{c|}{CIFAR-10} & \multicolumn{3}{c|}{CIFAR-100} & \multicolumn{3}{c}{ImageNet16-120} \\
 & Type & KT & SPR & $\text{nDCG}$ & KT & SPR & $\text{nDCG}$ & KT & SPR & $\text{nDCG}$ \\
 \hline
 \multicolumn{10}{c}{Simple rankings}\\
 \hline
FLOPs & S & 0.623 & 0.799 & 0.745 & 0.586 & 0.763 & 0.576 & 0.545 & 0.718 & 0.403 \\
GradNorm \cite{abdelfattah2021zero}& S  & 0.328 & 0.438 & 0.509 & 0.341 & 0.451 & 0.278 & 0.310 & 0.418 & 0.265 \\
GraSP \cite{abdelfattah2021zero, wang2020picking}& S  & 0.352 & 0.505 & 0.518 & 0.349 & 0.498 & 0.284 & 0.359 & 0.502 & 0.281 \\
SNIP \cite{abdelfattah2021zero, lee2018snip}& S  & 0.431 & 0.591 & 0.513 & 0.440 & 0.597 & 0.286 & 0.389 & 0.521 & 0.286 \\
SynFlow \cite{abdelfattah2021zero, tanaka2020pruning}& S  & 0.561 & 0.758 & 0.709 & 0.553 & 0.750 & 0.594 & 0.531 & 0.719 & 0.511 \\
Jacov \cite{abdelfattah2021zero}& S  & 0.616 & 0.800 & 0.540 & 0.639 & 0.820 & 0.402 & 0.602 & 0.779 & 0.356 \\
NASWOT \cite{mellor2021neural}& S  & 0.571 & 0.762 & 0.607 & 0.607 & 0.799 & 0.475 & 0.605 & 0.794 & 0.490 \\
ZenNAS \cite{lin2021zen}& S  & 0.102 & 0.103 & 0.120 & 0.079 & 0.072 & 0.115 & 0.091 & 0.109 & 0.073 \\
GradSign\dag~\cite{zhang2021gradsign}& S  & $\cdot$ & 0.765 & $\cdot$ & $\cdot$ & 0.793 & $\cdot$ & $\cdot$ & 0.783 & $\cdot$ \\
ZiCo \cite{li2023zico}& S  & 0.607 & 0.802 & 0.751 & 0.614 & 0.809 & 0.607 & 0.587 & 0.779 & 0.523 \\
TE-NAS \cite{chen2021neural}& A  & 0.536 & 0.722 & 0.602 & 0.537 & 0.723 & 0.327 & 0.523 & 0.709 & 0.330 \\
AZ-NAS \cite{lee2024az}& A  & 0.712 & 0.892 & 0.749 & 0.696 & 0.880 & 0.549 & 0.673 & 0.859 & 0.534 \\
No. of trainable layers ($\aleph$) & S & 0.626 & 0.767 & 0.671 & 0.646 & 0.787 & 0.525 & 0.623 & 0.764 & 0.497 \\
$\text{VKDNW}_{\text{single}}$ (ours)& S  & 0.618 & 0.815 & 0.751 & 0.634 & 0.829 & 0.617 & 0.622 & 0.814 & 0.608 \\
$\text{VKDNW}_{\text{agg}}$ (ours)& A  & \textbf{0.750} & \textbf{0.919} & \textbf{0.785} & \textbf{0.753} & \textbf{0.919} & \textbf{0.636} & \textbf{0.743} & \textbf{0.906} & \textbf{0.664} \\
 \hline
 \multicolumn{11}{c}{Model-driven rankings}\\
 \hline
GRAF \cite{kadlecova2024surprisingly} & A & 0.820 & 0.953 & 0.935 & 0.809 & 0.948 & 0.858 & 0.796 & 0.941 & 0.828 \\
$\text{VKDNW}_{m}$ (ours) & A & 0.647 & 0.831 & 0.750 & 0.636 & 0.821 & 0.602 & 0.611 & 0.798 & 0.575 \\
$\text{(VKDNW+ZCS)}_{m}$ (ours) & A & 0.840 & 0.963 & 0.922 & 0.834 & 0.960 & 0.884 & 0.830 & 0.958 & 0.843 \\
$\text{(VKDNW+ZCS+GRAF)}_{m}$ (ours) & A & \textbf{0.859} & \textbf{0.971} & \textbf{0.946} & \textbf{0.847} & \textbf{0.966} & \textbf{0.895} & \textbf{0.842} & \textbf{0.963} & \textbf{0.867}
\end{tabular}
\caption{Training-free NAS methods in the NAS-Bench-201~\cite{dong2020bench} search space, evaluated on three public datasets. Kendall's $\tau$ (KT), Spearman's $\rho$ (SPR) and Normalized Discounted Cumulative Gain ($\text{nDCG}$) are reported, results are averages of 5 independent runs. The Type column differentiates single (S) and aggregated (A) rankings. Results are reproduced with code published by their authors, except those marked\dag, where results from the original paper are taken. 
}
\label{tab:table1}
\end{table*}
Throughout previous works different approaches for ranking computation can be found. In many of the works the gradient with respect to the network weights is investigated (GradNorm, GraSP, SNIP, Synflow, \cite{tanaka2020pruning, wang2020picking, lee2018snip, abdelfattah2021zero}), i.e. leveraging the first order approximation of the network. In Jacov \cite{abdelfattah2021zero} correlations of the Jacobian matrices among various input samples are compared; in NASWOT \cite{mellor2021neural} the linear maps induced by data points are examined; Zen \cite{lin2021zen} uses the approximation of gradient with respect to featuremaps; GradSign \cite{zhang2021gradsign} compares the optimization landscape at the level individual training samples; or in ZiCo \cite{li2023zico} the gradients from multiple forward and backward passes preferring large magnitude and low variance.

Other methods aggregate multiple sources aiming at obtaining a better informed ranks: TE-NAS \cite{chen2021neural} uses both the number of linear regions \cite{hanin2019complexity, xiong2020number} and the condition number of Neural Tangent Kernel \cite{jacot2018neural, lee2019wide}. However, it is well known that the kernel computation is highly computationally demanding \cite{novak2022fast}. AZ-NAS \cite{lee2024az} assesses expressivity, trainability and progressivity via examination of feature distribution across all orientations and the Jacobian. However, despite AZ-NAS outperforming previous works in some metrics, it is worse than \cite{li2023zico} in key NAS-related aspects such as the cumulative gain (see \cref{subsec:ranking}).

Despite a tremendous effort of the community, it was shown that most of the zero-shot NAS methods perform worse than a simple proxy given just by FLOPs or \#params \cite{ning2021evaluating, white2022deeper}. Thus, there is still a gap for improvements also driven by the need of the ranking explainability that would have a satisfactory theoretical support.

Furthermore, \cite{kadlecova2024surprisingly} uses a model-driven ranking that however needs a set of networks for which the validation accuracies are known to train the model and the ability of this ranking to generalize as it is fitted on a specific network collection only is disputable.

Finally, let us discuss the current practice in the NAS methods comparison. Methods are compared either by means of correlations of their scores to accuracy or by reporting the accuracy of the top-ranked network \cite{lee2024az, li2023zico}. However, while the first is not tailor-made for the architecture search task and therefore does not assess the desired ranking properties (see Sec. \ref{subsec:ranking}), comparing performance just by the accuracy of a single network is too vulnerable e.g. to the random seed choice.  

\section{Experiments}
\label{sec:experiments}

\subsection{Ranking aggregation}
\label{subsec:ranking}

In addition to using the single ranking of \cref{eq:vkdnwrank}, we also experiment with multiple rankings  (as in~\cite{li2023zico, lee2024az, kadlecova2024surprisingly}) to order network architectures by their accuracy. We use two different options: non-linear and model-driven aggregation.

\paragraph{Non-linear aggregation.} The aggregation~\cite{lee2024az} uses multiplication to combine multiple ranks into a single one, which means a network is highly-ranked if and only if it is highly-ranked in \textit{all} subsidiary rankings, keeping their influence balanced. That is, denoting rankings $\rank_1,\dots, \rank_m$ for some $m\in\mathbb{N}$ we define the aggregated ranking
\begin{align}
    \rank_{\agg}(f):=\log{\Pi_{j=1}^m\rank_j(f)}
\end{align}
for each network $f$. This aggregation is therefore possible only in the context of a given network collection, such as in evolutionary search. 

In our case, $\text{VKDNW}_{\text{agg}}$ aggregates these five proxies:
\begin{itemize}
    \item $\text{VKDNW}_{\text{single}}$ (V) ranking (\cref{eq:vkdnwrank}),
    \item Jacov (J)~\cite{abdelfattah2021zero} measures the activations correlation when exposed to various inputs, 
    \item Expressivity (E)~\cite{lee2024az} assesses isotropy and uniformity of the features distribution across all orientations, 
    \item Trainability (T)~\cite{lee2024az} captures ability of the network to keep stable gradient propagation between the layers by inspecting the spectrum of the Jacobian matrix,
    \item FLOPs (F) is the number of FLOPs for one forward pass.
\end{itemize}

\paragraph{Model-driven aggregation.}
When accuracies of a sufficient number of architectures in the search space are known, 
model-driven aggregation can be used to train a regression model to combine individual rankings. The trained model is then used to predict accuracy for unseen networks in the same search space. We evaluate three different models:
\begin{itemize}
    \item $\text{VKDNW}_{m}$ where the eigenvalues $\lambda_k$ of the FIM matrix are used directly as features in companion with $\aleph$ (number of trainable layers) and FLOPs to allow a more complex proxy of the diversity of the eigenvalues and the network complexity than the simple entropy \cref{eq:vkdnw},
    \item $\text{(VKDNW+ZCS)}_{m}$ where we additionally include all other zero-cost scores available (see Table \ref{tab:table1}),
    \item $\text{(VKDNW+ZCS+GRAF)}_{m}$ where we add network graph features from \cite{kadlecova2024surprisingly}. 
\end{itemize}

\subsection{Results}
\label{subsec:results}

We have conducted experiments in the NAS-Bench-201 \cite{dong2020bench} and MobileNetV2 \cite{lin2021zen, sandler2018mobilenetv2} search spaces. To obtain easily comparable results, we used 64 randomly generated input images to compute our score as in \cite{lee2024az}. For the methods that rely on the knowledge of true labels, input data from the respective datasets were used.

\begin{table}
  \centering
  \small
  \begin{tabular}{@{}lcccc@{}}
    \toprule
    Method & FLOPs & Top-1 acc. & Type & Search cost \\
    &&&& (GPU days) \\
    \midrule
NASNet-B~\cite{zoph2018learning} & 488M & 72.8 & MS & 1800 \\
CARS-D~\cite{yang2020cars}  & 496M & 73.3 & MS & 0.4 \\
BN-NAS~\cite{chen2021bn}  & 470M & 75.7 & MS & 0.8 \\
OFA~\cite{cai2019once}  & 406M & 77.7 & OS & 50 \\
RLNAS~\cite{zhang2021neural}  & 473M & 75.6 & OS & - \\
DONNA~\cite{moons2021distilling}  & 501M & 78.0 & OS & 405 \\
\# Params & 451M & 63.5 & ZS & 0.02 \\
ZiCo~\cite{li2023zico}  & 448M & 78.1 & ZS & 0.4 \\
AZ-NAS~\cite{lee2024az} & 462M & 78.6 & ZS & 0.4 \\
$\text{VKDNW}_{\text{agg}}$ (ours) & 480M & \textbf{78.8} & ZS & 0.4 \\

    \bottomrule
  \end{tabular}
  \caption{Results on ImageNet-1K \cite{deng2009imagenet} in the MobileNetV2 search space, the size of the model is constrained to $\approx$450M FLOPS. 
  }
  \label{tab:mobileNertQuantiative}
\end{table}

\paragraph{NAS-Bench-201.} The dataset consists of 15,625 networks for which validation accuracies for CIFAR-10, CIFAR-100 \cite{krizhevsky2009learning} and ImageNet16-120 \cite{chrabaszcz2017downsampled} after training for 200 epochs are provided. The networks are characterized by unique cell structures comprising of several types of operation choices. As one of the possible choices is zero operation, it's possible that some computation edges don't receive any input or cannot propagate their results to the output, leading to same computation graphs for different architectures, thus duplicating networks. Following the practice from NAS-Bench-101 \cite{ying2019bench}, we report results on 9,445 unique structures (also in \cite{mehrotra2021bench, kadlecova2024surprisingly}) and we refer the reader to Supplementary material for results on all networks. 
We measure proxies performance via Kendall’s $\tau$ (KT) and Spearman's $\rho$ (SPR) correlations with validation accuracies together with Normalized Discounted Cumulative Gain ($\text{nDCG}_{1000}$, we write nDCG for short) from \cref{sec:evaluationmetrics}, averaged over 5 independent runs. For the model-driven aggregation, we trained a random forest model on 1024 networks for 100 iterations, and used the rest for testing.

\begin{figure}
    \centering    
    \includegraphics[width=\columnwidth,trim=60 60 60 60,clip]{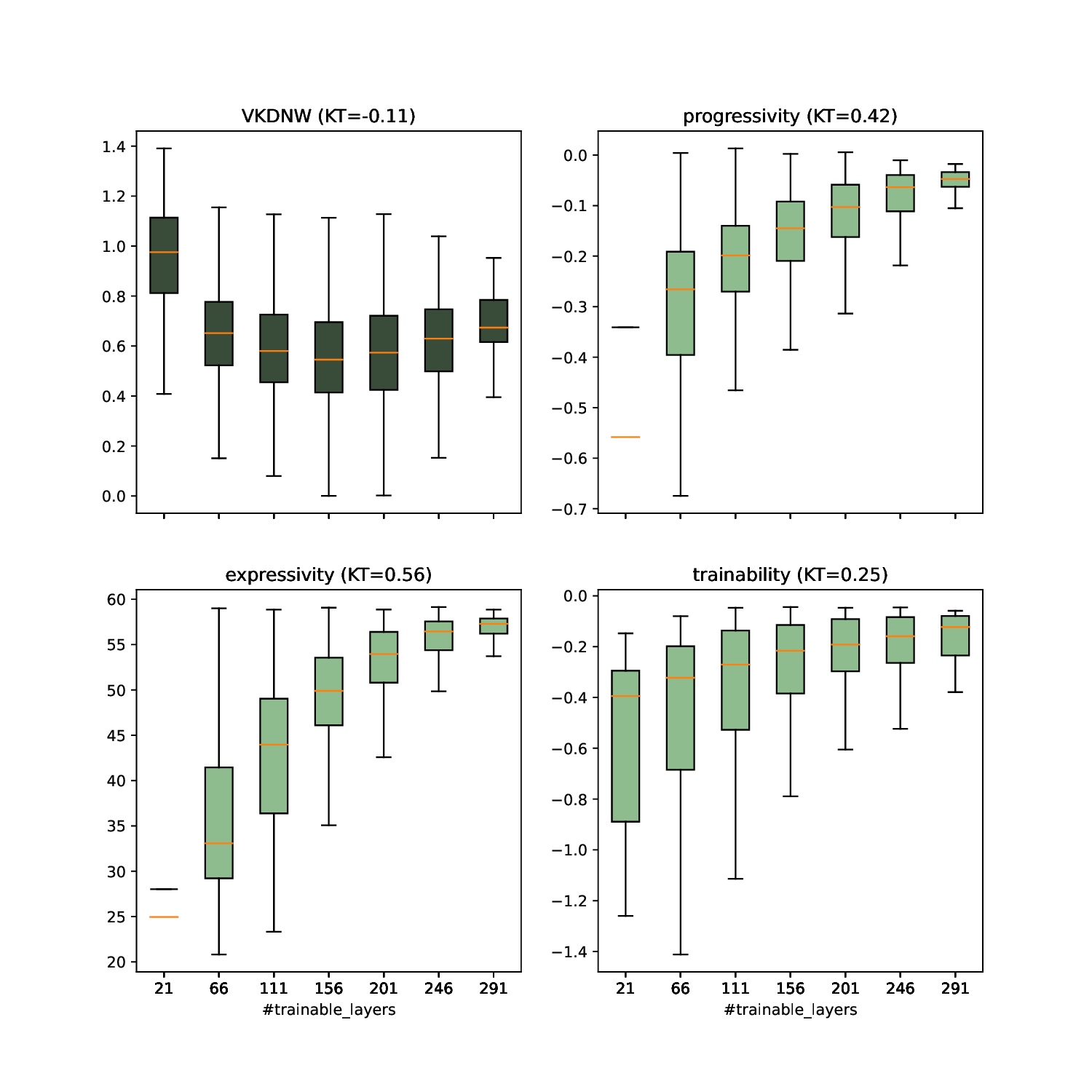}
    \caption{Components of AZ-NAS \cite{lee2024az} and our VKDNW are compared w.r.t. correlation with $\aleph$ (number of trainable layers), in the NAS-Bench-201 search space \cite{dong2020bench} on ImageNet16-120 \cite{chrabaszcz2017downsampled} dataset. Our VKDNW proxy has the lowest correlation, ie. is the most invariant to the size of the model.\vspace{-15pt}}
    
    \label{fig:boxplot_per_trainable_layers}
\end{figure}
In \Cref{tab:table1}, we compare our method with existing NAS methods and show  that $\text{VKDNW}_{\text{agg}}$ outperforms all methods with a significant margin in all three metrics. The benefit of our method is two-fold: a) it identifies the best networks due to its high performance in $\text{nDCG}$, which is the desirable property in NAS b) the ranking is consistent across the whole network search space due to high KT and SPR correlations. Note that we can only make these observations by combining standard correlation metrics and our newly proposed $\text{nDCG}$. We also show that $\text{VKDNW}_{\text{single}}$ outperforms all other single-rank proxies in all metrics on ImageNet16-120, in SPR on all three datasets, while being among the highest on \mbox{CIFAR-10 and CIFAR-100}. 

\paragraph{MobileNetV2.}
We search for the best network configuration in the MobileNetV2 space~\cite{sandler2018mobilenetv2}, while constraining the model size to approximately 450M FLOPs. We ran 100,000 iterations of the evolutionary search algorithm~\cite{lee2024az} for approximately 10 hours, meaning 100,000 different architectures were evaluated. We then took the best network from the search and trained it for 480 epochs on  ImageNet-1K~\cite{deng2009imagenet}, using the same hyper-parameter setting as in \cite{lee2024az, li2023zico}. The final training of the model took 7 days on 8xNVidia A100 GPUs.

As seen in \Cref{tab:mobileNertQuantiative}, our method $\text{VKDNW}_{\text{agg}}$ outperforms all prior approaches -- even train-based approaches (denoted MS and OS) that incur much higher computational costs for the search.

\begin{table}
    \centering
\small
\begin{tabular}{ccccc|ccc}
\hline
V & J & E & T & F & (KT) & (SPR) & $\text{nDCG}$ \\
 \hline
 \checkmark & & & & & \textbf{0.622} & \textbf{0.814} & \textbf{0.608} \\
 & \checkmark & & & & 0.603 & 0.781 & 0.339 \\
 & & \checkmark & & & 0.588 & 0.779 & 0.274 \\
& & & \checkmark & & 0.353 & 0.517 & 0.233 \\
& & & & \checkmark & 0.545 & 0.718 & 0.403 \\
\hline
\checkmark & \checkmark & \checkmark & \checkmark & & 0.717 & 0.891 & 0.623 \\
\checkmark & \checkmark & \checkmark & & \checkmark & 0.706 & 0.871 & 0.553 \\
\checkmark & \checkmark & & \checkmark & \checkmark & \textbf{0.736} & \textbf{0.905} & \textbf{0.695} \\
\checkmark & & \checkmark & \checkmark & \checkmark & 0.722 & 0.896 & 0.658 \\
& \checkmark & \checkmark & \checkmark & \checkmark & 0.735 & 0.901 & 0.646 \\
\hline
\checkmark & \checkmark & \checkmark & \checkmark & \checkmark & \textbf{0.743} & \textbf{0.906} & \textbf{0.664} \\
\hline 
\end{tabular} 
\caption{Components of $\text{VKDNW}_{\text{agg}}$ rank with non-linear aggregation. Consistency is shown with respect to Kendall's $\tau$ (KT), Spearman's $\rho$ (SPR) and Normalized Discounted Cumulative Gain ($\text{nDCG}$) with $P=1000$ on ImageNet16-120 image dataset \cite{chrabaszcz2017downsampled}. Here V, J, E, T and F stand for VKDNW, Jacov, expressivity, trainability and FLOPs respectively (see Sec. \ref{subsec:ranking}). Table of all combinations can be found in the Supplementary material. 
}
\label{tab:ablationComponentsSmall}
\end{table}

\subsection{Ablations}
\label{subsec:ablation}

We ablate our method in the NAS-Bench-201 search space \cite{dong2020bench} on ImageNet16-120 \cite{chrabaszcz2017downsampled} validation accuracies.
\paragraph{Orthogonality of VKDNW.} Our score VKDNW has a desirable property that it is based on information orthogonal to the size of the network: in Figure \ref{fig:boxplot_per_trainable_layers} we can see that unlike previous work, VKDNW is not correlated with the network size measured by $\aleph$ (number of trainable layers). We believe this property is key when stepping into much larger search spaces, however components of the previous state-of-the-art method AZ-NAS lack such a property (also \cref{fig:boxplot_per_trainable_layers} and \cref{tab:kendall_by_trainable_layers}). We conjecture that improvement in the key metrics by VKDNW is caused by this orthogonality feature. In Table \ref{tab:table1} we also provide comparison to previous model-driven method and show that adding our feature significantly improves the ranking in all considered metrics.
\begin{table}
  \centering
  \small
  \begin{tabular}{l|ccc}
  \hline
  & \multicolumn{3}{c}{(KT)}\\
  & CIFAR-10 & CIFAR-100 & ImageNet16-120 \\  
\hline
VKDNW & \applygradient{0.041}{0}{1}0.041 & \applygradient{0.090}{0}{1}-0.090 & \applygradient{0.107}{0}{1}-0.107 \\
trainability & \applygradient{0.220}{0}{1}0.220 & \applygradient{0.220}{0}{1}0.220 & \applygradient{0.253}{0}{1}0.253 \\
expressivity & \applygradient{0.539}{0}{1}0.539 & \applygradient{0.539}{0}{1}0.539 & \applygradient{0.539}{0}{1}0.539 \\
progressivity & \applygradient{0.398}{0}{1}0.398 & \applygradient{0.398}{0}{1}0.398 & \applygradient{0.385}{0}{1}0.385 \\ \hline
  \end{tabular}
  \caption{Kendall's $\tau$ (KT) of $\aleph$ (number of trainable layers) with components of AZ-NAS compared to our new method VKDNW on three datasets on NAS-Bench-201 search space \cite{dong2020bench} and multiple image datasets.
  }
    \label{tab:kendall_by_trainable_layers}
\end{table}

\paragraph{Components of aggregated rank.} Our single-rank variant $\text{VKDNW}_{\text{single}}$ is the strongest component of the aggregated $\text{VKDNW}_{\text{agg}}$ as can be seen in Table \ref{tab:ablationComponentsSmall} where it outperforms rest of the rankings with especially large margin in nDCG, thus, it strongly discriminates high-accuracy networks compared to others. 

\begin{table}
  \centering
  \small
  \begin{tabular}{c|ccc|ccc}
  \hline
Batch & \multicolumn{3}{c|}{Random Input} & \multicolumn{3}{c}{Real Input} \\ \cline{2-7}
 Size & KT & SPR & $\text{nDCG}$ & KT & SPR & $\text{nDCG}$ \\
\hline
8 & 0.615 & 0.808 & 0.617 & 0.614 & 0.805 & 0.599 \\
16 & 0.614 & 0.808 & 0.604 & 0.608 & 0.800 & 0.595 \\
32 & 0.619 & 0.812 & 0.611 & 0.612 & 0.804 & 0.594 \\
64 & 0.621 & 0.814 & 0.617 & 0.614 & 0.806 & 0.595 \\
128 & 0.621 & 0.814 & 0.616 & 0.615 & 0.806 & 0.591 \\
\hline
  \end{tabular}
  \caption{Our method VKDNW evaluated for different batch sizes using either randomly generated input data or real images with respect to Kendall's $\tau$ (KT), Spearman's $\rho$ (SPR) and Normalized Discounted Cumulative Gain with $P=1000$ (nDCG).
  }
    \label{tab:real_vs_random}
\end{table}

\paragraph{Random or real input.} As our method is computed using generated random data (white noise) it does not rely on any real dataset and is therefore applicable also in situations where no reliable data are available. Table \ref{tab:real_vs_random} shows that by relying just on random data we do not lose any performance with respect to all key metrics. Moreover, the performance remains relatively stable across different batch sizes. We chose batch size 64 as larger batch size does not bring better results.

Due to lack of space, we kindly refer the reader to Supplementary material for further ablations.

\vspace{-0.5em}
\section{Conclusion}
\label{sec:conclusion}

We proposed a new training-free NAS proxy called \textit{Variance of Knowledge of Deep Network Weights (VKDNW)}. The method has strong theoretical support and achieved state-of-the-art results on three public datasets and in two search spaces. 

Based on throughout evaluation and comparison to other existing approaches, it has been shown that it provides an information orthogonal to the network size leading to a zero-cost ranking were contribution of the network size and architecture feasibility are separated. We have also shown that previously used correlation metrics for proxy evaluations do not sufficiently assess the key ability to discriminate top networks and to address this problem we have proposed a new evaluation metric and re-evaluated previous methods with the new metric. 
%

{\small
    \bibliographystyle{ieeenat_fullname}
    \bibliography{main}
}

\clearpage
\twocolumn[
\centering
\Large
\vspace{0.5em}Supplementary Material \\
\vspace{1.0em}
]

In the supplement material, we elaborate our formal arguments and provide additional results and ablations.

\section{Fisher Information}
\label{suppl:fim}

In this section, we provide a more detailed inspection on the Fisher Information matrix (FIM) in the context of neural networks as an extension of \cref{subsec:fim}. The \Cref{subsec:cramer-rao} extends some results from the main paper, the \Cref{subsec:ngd} provides another motivation on why the FIM should be considered as a tool for analysis of neural networks and finally \Cref{subsec:montecarlo} summarizes terminology issues within the community. 

\subsection{Cramér-Rao bound}
\label{subsec:cramer-rao}
Consider the same setting as in \cref{subsec:fim}, that is a deep network $f$ is given with the unknown (deterministic) weight vector $\theta\in\mathbb{R}^p$ for some parameters $p\in\mathbb{N}$, whose estimation is the subject of our interest. Take any estimator $\Hat{\theta}_n$ that is computed from $n$ independently-drawn input images and which is unbiased, i.e.
\begin{align}
    \mathbb{E}\Hat{\theta}_n=\theta.
    \label{supp:unbiased}
\end{align}
Then we have the lower bound for the variance matrix of $\Hat{\theta}_n$ given as
\begin{align}
    \text{Var}\left(\Hat{\theta}_n\right)\geq \frac{1}{n}F^{-1}(\theta),
    \label{supp:cramer_rao}
\end{align}
which in turn gives also an estimate for the diagonal elements
\begin{align}
    \left(\text{Var}\left(\Hat{\theta}_n\right)\right)_{jj}\geq \frac{1}{n}\left(F^{-1}(\theta)\right)_{jj},
    \label{supp:cramer_rao_diag}
\end{align}
where we use the standard notation $A_{ij}$ for the entry at the position $i,j$ for any matrix $A$.

We now show the relation between \eqref{supp:cramer_rao} and mean-square error of the weight estimator. First, consider a weight $\theta(j)$ for some $j\in\lbrace 1,\dots, p\rbrace$. Then we shall write $\theta(j)=e_j^T\theta$, where $e_j$ is the $j$th unit vector consisting only of zeros and single one at the $j$th position: $e_j=(0,\dots,1,\dots, 0)$. Now recall that that if a random $d$-dimensional vector $X$ has a variance matrix $\text{Var}\left( X\right)$ and $e\in\mathbb{R}^d$ then the linear combination $e^TX$ has the variance
\begin{align}
    \text{Var}\left(e^TX\right)=e\text{Var}\left( X\right)e^T,
\end{align}
from which it easily follows that the variance of the random scalar $\Hat{\theta}_n(j)$ is
\begin{align}
    \text{Var}\left( \Hat{\theta}_n(j)\right)=&\text{Var}\left( e_j^T\Hat{\theta}_n\right)\nonumber\\
    =&e_j\text{Var}\left( \Hat{\theta}_n\right)e_j^T\nonumber\\
    =&\left(\text{Var}\left( \Hat{\theta}_n\right)\right)_{jj}.
    \label{supp:variance_diagonal}
\end{align}
Next, we use the bias-variance decomposition of the mean-square error: if $\Hat{X}$ is an estimator of an unknown scalar value $X\in\mathbb{R}$, then
\begin{align}
    \mathbb{E}\left(\Hat{X}-X\right)^2=&\mathbb{E}\left(\Hat{X}-\mathbb{E}\Hat{X}+\mathbb{E}\Hat{X}-X\right)^2\nonumber\\
    =&\mathbb{E}\left(\Hat{X}-\mathbb{E}\Hat{X}\right)^2+2\mathbb{E}\left(\Hat{X}-\mathbb{E}\Hat{X}\right)\mathbb{E}\left(\Hat{X}-X\right)\nonumber\\
    +&\mathbb{E}\left(\mathbb{E}\Hat{X}-X\right)^2\nonumber\\
    =&\text{Var}\Hat{X}+\left(\mathbb{E}\Hat{X}-X\right)^2\nonumber\\
    =&\text{Var}\Hat{X}+\left(\text{Bias}\hat{X}\right)^2.
    \label{supp:tradeoff}
\end{align}
Combining \eqref{supp:unbiased} with \eqref{supp:variance_diagonal}, \eqref{supp:tradeoff} and the Cramér-Rao bound \eqref{supp:cramer_rao} we obtain
\begin{align}
    \mathbb{E}\left( \Hat{\theta}_n(j)-\theta(j)\right)^2=&\text{Var}\left({\Hat{\theta}_n(j)}\right)+\left(\text{Bias}\left(\Hat{\theta}_n(j)\right)\right)^2\nonumber\\
    =&\text{Var}\left({\Hat{\theta}_n(j)}\right)\nonumber\\
    =&\left(\text{Var}\left( \Hat{\theta}_n\right)\right)_{jj}\nonumber\\
    \geq&\frac{1}{n}\left(F^{-1}(\theta)\right)_{jj}.
\end{align}
Summing now over all indices $j$ we finally obtain a lower bound of for the mean-square error of the entire weight vector $\Hat{\theta}_n$
\begin{align}
    \mathbb{E}\|\theta_1-\theta_2\|^2=&\sum_{j=1}^p\mathbb{E}\left( \Hat{\theta}_n(j)-\theta(j)\right)^2\nonumber\\
    \geq&\frac{1}{n}\sum_{j=1}^p\left(F^{-1}(\theta)\right)_{jj}.
    \label{supp:mean_square}
\end{align}
The quantity on the right-hand-side of \eqref{supp:mean_square} is the trace of the matrix $F^{-1}(\theta)$ and it coincides with sum of the eigenvalues of $F^{-1}(\theta)$, which are just the reciprocals of the eigenvalues of $F(\theta)$ \cite{leon2006linear}. 

We have shown that eigenvalues of the FIM determine the least-possible mean-square error for any unbiased estimator of the network weight vector $\theta$ and its components. The case of a biased estimator is more delicate and we kindly refer the reader to \cite{frieden2010exploratory}.

\subsection{Natural Gradient Descent}
\label{subsec:ngd}
Natural Gradient Descent (see \cite{martens2020new}) is an improvement of the classical Stochastic Gradient Descent that is proven to have faster and more stable convergence, but for the price of significantly increased computation costs. In Natural Gradient Descent, the weight updates are governed by a transformed loss gradient as
\begin{align}
    \theta_{n+1}=\theta_n-F^{-1}(\theta_n)\nabla_\theta \mathcal{L}(\theta_n),
\end{align}
where $F^{-1}(\theta)$ is the inverse of the FIM and $\mathcal{L}(\theta)$ is the loss function. 

We also take the steepest descent direction of the loss function, but now we do not measure the distance in the space of weights by means of the Euclidean distance but we adjust the curvature by measuring the KL-divergence of the output distributions. In other words, Natural Gradient Descent is just what happens to Stochastic Gradient Descent if we say that two weight vectors $\theta_1, \theta_2$ are close to each other if
\begin{align}
    D_{KL}&(\sigma_{\theta_1}(\cdot|x), \sigma_{\theta_2}(\cdot|x))
\end{align}
is small, in contrast to the usual case when we consider $\|\theta_1-\theta_2\|$ instead.  And again similarly as above, after inspecting the eigenvalues of the FIM we can conclude that the more different the eigenvalues are the more difficult is to train the model as the weight updates are much larger in some directions than in others. 

Even though we later use the FIM in the applications where the networks have been trained with the classical Stochastic Gradient Descent, the curvature given by the FIM still provides a valuable information -- if the network is more difficult to train using Natural Gradient Descent, it's unlikely that when using a simpler optimisation method, the network would yield stronger performance after training.

\subsection{Monte Carlo estimation of the Fisher Information Matrix (FIM)}
\label{subsec:montecarlo}

We now follow \cite{kunstner2019limitations} and outline details of the common misconception in the terminology within the community that might lead to incorrect estimation of the FIM. 

In our setting, the FIM is given as
\begin{align}
\label{supp:fim}
F(\theta)\coloneq\mathbb{E}\left[\nabla_{\theta}\sigma_\theta(c\,|\,x)\,\nabla_{\theta}\sigma_\theta(c\,|\,x)^T\right]\in\mathbb{R}^{p\times p},
\end{align}
where the expectation $\mathbb{E}$ is taken with respect to the join distribution of the image-label pair $(x,c)$. Recall that the joint distribution can be decomposed into the prior distribution for $x$, usually unknown, and the conditional distribution distribution for the label $c$ given $x$
\begin{align}
    \sigma_\theta(c\,|\,x)=\frac{\exp\left({\Psi_c(x, \theta)}\right)}{\sum_{d=1}^C\exp\left({\Psi_d(x, \theta)}\right)}, \quad c=1,\dots, C
\end{align}
where $\Psi(x,\theta)\in\mathbb{R}^C=\left( \Psi_1(x,\theta),\dots ,\Psi_C(x,\theta)\right)$ is the network output (logits) given the weight vector $\theta\in\mathbb{R}^p$, i.e. the network weights. We might deal with missing information on the prior distribution of $x$ by simply replacing it with the empirical distribution given by independently drawn examples $x_1,\dots, x_n$ which then yields a Monte Carlo estimate
\begin{align}
\Hat{F}(\theta)&\coloneq\frac{1}{n}\sum_{n=1}^N\mathbb{E}_{\sigma_\theta}\left[\nabla_{\theta}\sigma_\theta(c\,|\,x_n)\,\nabla_{\theta}\sigma_\theta(c|\,x_n)^T\right]
    \label{supp:fim_empirical}
\end{align}
where $\mathbb{E}_{\sigma_\theta}$ now denotes the expectation with respect to the model prediction $\sigma_\theta$, which is in the statistical community denoted as the empirical FIM. From the strong law of large numbers it follows that the empirical FIM converges to the FIM almost surely as the number of samples $n$ tends to infinity. Therefore, it is reasonable to replace \eqref{supp:fim} in the applications by \eqref{supp:fim_empirical}.

However, in some methods (see \cite{kunstner2019limitations} and references therein) the expectation in \eqref{supp:fim_empirical} with respect to the model prediction $\sigma_\theta$ is often replaced by the empirical distribution $\sigma$ of the labels given the images which leads to a different definition
\begin{align}
G(\theta)&\coloneq\frac{1}{n}\sum_{n=1}^N\mathbb{E}_{\sigma}\left[\nabla_{\theta}\sigma_\theta(c\,|\,x_n)\,\nabla_{\theta}\sigma_\theta(c|\,x_n)^T\right]\nonumber\\
&=\frac{1}{n}\sum_{n=1}^N\left[\nabla_{\theta}\sigma_\theta(c_n\,|\,x_n)\,\nabla_{\theta}\sigma_\theta(c_n|\,x_n)^T\right],
    \label{supp:fim_empirical_wrong}
\end{align}
where now $(x_j\,c_j)$ are the observed image-label pairs. The difference between \eqref{supp:fim_empirical} and \eqref{supp:fim_empirical_wrong} is that in the former we sum the multiplied gradients over all categories $c$ weighted by the network-predicted probability, while in the later we use only single class as if the network correctly classified the sample with zero error. At the initialization stage, the network prediction is however far from the ground-truth distribution and therefore $G(\theta)$ is indeed very different from the Monte Carlo approximation $\Hat{F}(\theta)$ (and also from the FIM $F(\theta)$ itself). In our method we used $\Hat{F}(\theta)$.

\section{Experiments}

\begin{table*}[t]
    \centering    
\begin{tabular}{lc|ccc|ccc|ccc}
\hline
 & & \multicolumn{3}{c|}{CIFAR-10} & \multicolumn{3}{c|}{CIFAR-100} & \multicolumn{3}{c}{ImageNet16-120} \\
 & Type & KT & SPR & $\text{nDCG}$ & KT & SPR & $\text{nDCG}$ & KT & SPR & $\text{nDCG}$ \\
 \hline
 \multicolumn{10}{c}{Simple rankings}\\
 \hline
FLOPs & S & 0.578 & 0.753 & 0.729 & 0.551 & 0.727 & 0.565 & 0.517 & 0.691 & 0.386 \\
GradNorm \cite{abdelfattah2021zero}& S & 0.356 & 0.483 & 0.407 & 0.359 & 0.489 & 0.202 & 0.322 & 0.441 & 0.192 \\
GraSP \cite{abdelfattah2021zero, wang2020picking}& S & 0.315 & 0.454 & 0.439 & 0.322 & 0.461 & 0.224 & 0.333 & 0.470 & 0.207 \\
SNIP \cite{abdelfattah2021zero, lee2018snip} & S & 0.454 & 0.615 & 0.433 & 0.462 & 0.620 & 0.221 & 0.403 & 0.539 & 0.212 \\
SynFlow \cite{abdelfattah2021zero, tanaka2020pruning}& S & 0.571 & 0.769 & 0.691 & 0.565 & 0.761 & 0.584 & 0.555 & 0.747 & 0.504 \\
Jacov \cite{abdelfattah2021zero}& S & 0.545 & 0.712 & 0.362 & 0.554 & 0.720 & 0.249 & 0.537 & 0.701 & 0.240 \\
NASWOT \cite{mellor2021neural}& S & 0.556 & 0.742 & 0.572 & 0.579 & 0.768 & 0.449 & 0.583 & 0.768 & 0.459 \\
ZenNAS \cite{lin2021zen}& S & 0.244 & 0.321 & 0.110 & 0.232 & 0.300 & 0.110 & 0.250 & 0.344 & 0.065 \\
GradSign\dag~\cite{zhang2021gradsign}& S & $\cdot$ & 0.765 & $\cdot$ & $\cdot$ & 0.793 & $\cdot$ & $\cdot$ & 0.783 & $\cdot$ \\
ZiCo \cite{li2023zico}& S & 0.590 & 0.785 & 0.732 & 0.600 & 0.794 & 0.597 & 0.594 & 0.787 & 0.516 \\
TE-NAS \cite{chen2021neural}& A & 0.489 & 0.676 & 0.481 & 0.481 & 0.664 & 0.214 & 0.459 & 0.641 & 0.143 \\
AZ-NAS \cite{lee2024az}& A & 0.739 & \textbf{0.912} & 0.702 & 0.722 & 0.899 & 0.473 & 0.694 & 0.876 & 0.482 \\\
No. of trainable layers ($\aleph$) & S & 0.580 & 0.723 & 0.631 & 0.594 & 0.737 & 0.491 & 0.574 & 0.716 & 0.483 \\
$\text{VKDNW}_{\text{single}}$ (ours)& S & 0.606 & 0.800 & 0.724
 & 0.613 & 0.807 & 0.592 & 0.605 & 0.795 & 0.583 \\
$\text{VKDNW}_{\text{agg}}$ (ours)& A & \textbf{0.740} & 0.911 & \textbf{0.743} & \textbf{0.736} & \textbf{0.906} & \textbf{0.578} & \textbf{0.723} & \textbf{0.893} & \textbf{0.614} \\
 \hline
 \multicolumn{11}{c}{Model-driven rankings}\\
 \hline
GRAF \cite{kadlecova2024surprisingly} & A & 0.832 & 0.957 & 0.921 & 0.818 & 0.952 & 0.859 & 0.812 & 0.946 & 0.832 \\
$\text{VKDNW}_{m}$ (ours) & A & 0.682 & 0.864 & 0.757 & 0.655 & 0.840 & 0.573 & 0.642 & 0.826 & 0.506 \\
$\text{(VKDNW+ZCS)}_{m}$ (ours) & A & 0.865 & 0.973 & 0.906 & 0.859 & 0.970 & 0.861 & 0.863 & 0.971 & 0.828 \\
$\text{(VKDNW+ZCS+GRAF)}_{m}$ (ours) & A & \textbf{0.878} & \textbf{0.978} & \textbf{0.927} & \textbf{0.869} & \textbf{0.975} & \textbf{0.871} & \textbf{0.874} & \textbf{0.975} & \textbf{0.856} \\
\end{tabular}
\caption{Training-free NAS methods in the NAS-Bench-201~\cite{dong2020bench} search space with inaccessible nodes (see discussion in Sec. \ref{subsec:results}, evaluated on three public datasets. Kendall's $\tau$ (KT), Spearman's $\rho$ (SPR) and Normalized Discounted Cumulative Gain ($\text{nDCG}$) are reported, results are averages of 5 independent runs. The Type column differentiates single (S) and aggregated (A) rankings. Results are reproduced with code published by their authors, except those marked\dag, where results from the original paper are taken. 
}
\label{tab:table1_unfiltered}
\end{table*}

\paragraph{NAS-Bench-201.}

In \cref{tab:table1}, we provide results for the NAS-Bench-201 architecture search space where we adopted the practice of NAS-Bench-101 \cite{ying2019bench, kadlecova2024surprisingly, mehrotra2021bench} where only unique graph structures are considered. As we described in  \cref{sec:experiments}, the entire search space in NAS-Bench-201 contains also networks where some computation edges don't receive any input or their output cannot be propagated through the network due to the existence of zero operation nodes. By filtering these networks, the number of architectures drops from 15,625 to 9,445 unique architectures. We argue that this is indeed good practice as networks with unreachable parameters should not be used in practice as the energy costs rise without improved performance. Moreover, many of the ranking scores (such as FLOPs or \#params) do not make sense in such cases, because parameters/operations are not used in network output yet they are still included in these metrics. 

For the sake of completeness however, in \Cref{tab:table1_unfiltered} we provide results of our experiments on full NAS-Bench-201 search space, using all 15,625 architectures. We can see that both $\text{VKDNW}_{\text{single}}$ and $\text{VKDNW}_{\text{agg}}$ outperform all other simple rankings in all metrics on CIFAR-100 and ImageNet16-120. On CIFAR-10 dataset AZ-NAS\cite{lee2024az} achieves similar Kendall's $\tau$ and Spearman's $\rho$ correlations as $\text{VKDNW}_{\text{agg}}$, however $\text{VKDNW}_{\text{agg}}$ leads in nDCG with a considerable margin.

\paragraph{MobileNetV2.}
In this experiment, we search for the best network configuration in the MobileNetV2 space~\cite{sandler2018mobilenetv2}. The search space is much larger as it consists of different architectures with inverted residual blocks, where depth, width, and expansion ratio of the blocks is altered. We constrained the model size to approximately 450M FLOPs and number of layers to 14. We adapt the evolutionary search algorithm \cite{lee2024az} by replacing the objective function in the search algorithm with our $\text{VKDNW}_{\text{agg}}$.
We then ran 100,000 iterations of the algorithm, always keeping top 1,024 best architectures, measured by $\text{VKDNW}_{\text{agg}}$. In each iteration, one mutation operation randomly changes one element in one of the top 1,024 architectures, and the newly created architecture is again ranked using $\text{VKDNW}_{\text{agg}}$. As a result, 100,000 iterations of architecture evaluations were made in the search process, leaving us with a shortlist of 1,024 architectures in the end.

Out of these final 1,024 architectures, we then again picked the one with the highest $\text{VKDNW}_{\text{agg}}$ rank and trained it for 480 epochs on ImageNet-1K~\cite{deng2009imagenet} in the same teacher-student setting as \cite{li2023zico, lee2024az}. We used vanilla SGD optimizer with LR=0.2 and single-cycle cosine learning rate schedule. The final training of the model took 7 days on 8xNVidia A100 GPUs.

\section{Ablations}

\begin{table}
  \centering
  \small
  \begin{tabular}{c|ccc}
  \toprule
 FIM Dimension & KT & SPR & $\text{nDCG}$ \\
\hline
8 & 0.590 & 0.782 & 0.579 \\
16 & 0.619 & 0.810 & 0.592 \\
32 & 0.621 & 0.813 & 0.600 \\
64 & 0.621 & 0.814 & 0.607 \\
128 & 0.619 & 0.812 & 0.611 \\
256 & 0.619 & 0.812 & 0.606 \\
\bottomrule
  \end{tabular}
  \caption{Our method $\text{VKDNW}_{\text{single}}$ evaluated for different FIM (see Sec \ref{subsec:fim}) sizes with respect to Kendall's $\tau$ (KT), Spearman's $\rho$ (SPR) and Normalized Discounted Cumulative Gain with $P=1000$ (nDCG).}
    \label{tab:fim_dimension}
\end{table}

\paragraph{Fisher Information matrix size.}
In \cref{tab:fim_dimension}, we evaluate our method with varying number of trainable layers considered in the computation of the FIM (see \cref{eq:fim}). We can see that initial 16 layers of the network already carry enough information, even comparable to when we use 256 layers. In our method, we set this parameter to 128 to maximize for (nDCG) while keeping other metrics high.

\begin{table}
  \centering
  \small
  \begin{tabular}{c|ccc}
  \toprule
\multicolumn{4}{c}{One weight per layer} \\
\hline
Policy & KT & SPR & $\text{nDCG}$ \\
\hline
random & 0.618 & 0.811 & 0.586 \\
0 & 0.622 & 0.814 & 0.608 \\
0.2 & 0.622 & 0.815 & 0.602 \\
0.4 & 0.625 & 0.817 & 0.606 \\
0.6 & 0.623 & 0.814 & 0.598 \\
0.8 & 0.622 & 0.814 & 0.609 \\
1 & 0.621 & 0.813 & 0.608 \\
\hline
\multicolumn{4}{c}{Multiple weights per layer}\\
\hline
No. weights & KT & SPR & $\text{nDCG}$ \\
\hline
1 & 0.621 & 0.813 & 0.600 \\
2 & 0.634 & 0.824 & 0.608 \\
4 & 0.626 & 0.817 & 0.600 \\
8 & 0.605 & 0.797 & 0.594 \\
\bottomrule
  \end{tabular}
  \caption{Our method $\text{VKDNW}_{\text{single}}$ evaluated for different parameter sampling policies within each trainable layer. Two types of sampling methods are presented. In \textit{One weight per layer} we take 128 initial network layers and either sample one weight per layer randomly or we take for $p=0, 0.2,\dots, 1$ the $p$th index relative within the weight vector. In \textit{Multiple weights per layer} we take 32 initial network layers and sample uniformly $k$ weights for $k=1,2,4,8$. We evaluate Kendall's $\tau$ (KT), Spearman's $\rho$ (SPR) and Normalized Discounted Cumulative Gain with $P=1000$ (nDCG).}
    \label{tab:fim_index_choice}
\end{table}
\paragraph{Parameter sampling policy.}
To make the dimension of the FIM feasible for computation of eigenvalues, we use only a small portion of the network weights. More specifically, instead of taking the full matrix of dimension $p$ (number of trainable parameters), we only sample one weight from each trainable layer from the first 128 layers and compute the FIM as if the network did not have any other parameters. In \Cref{tab:fim_index_choice}, we compare performance of our method $\text{VKDNW}_{\text{single}}$ as we vary the number of weights per layer and their sampling policy. We can see that the performance as measured by (nDCG) is roughly the same when taking anything between one and four weights per layer, and then starts to slowly decrease with a higher number of weights per layer. Secondly, our method is robust against choice of the policy as the performance for the case of one weight per layer with changing position of the weight within each layer does not change significantly. To further show that we do not lose any performance when dealing only with limited number of initial layers, we show in \cref{tab:fim_dimension} that our method is also robust against change of number of considered layers (the highest number of layers we tested was 256 as the number of larger networks in NAS-Bench-201 is small).

\begin{figure}
    \centering    
    \includegraphics[width=\columnwidth,trim=60 60 60 60,clip]{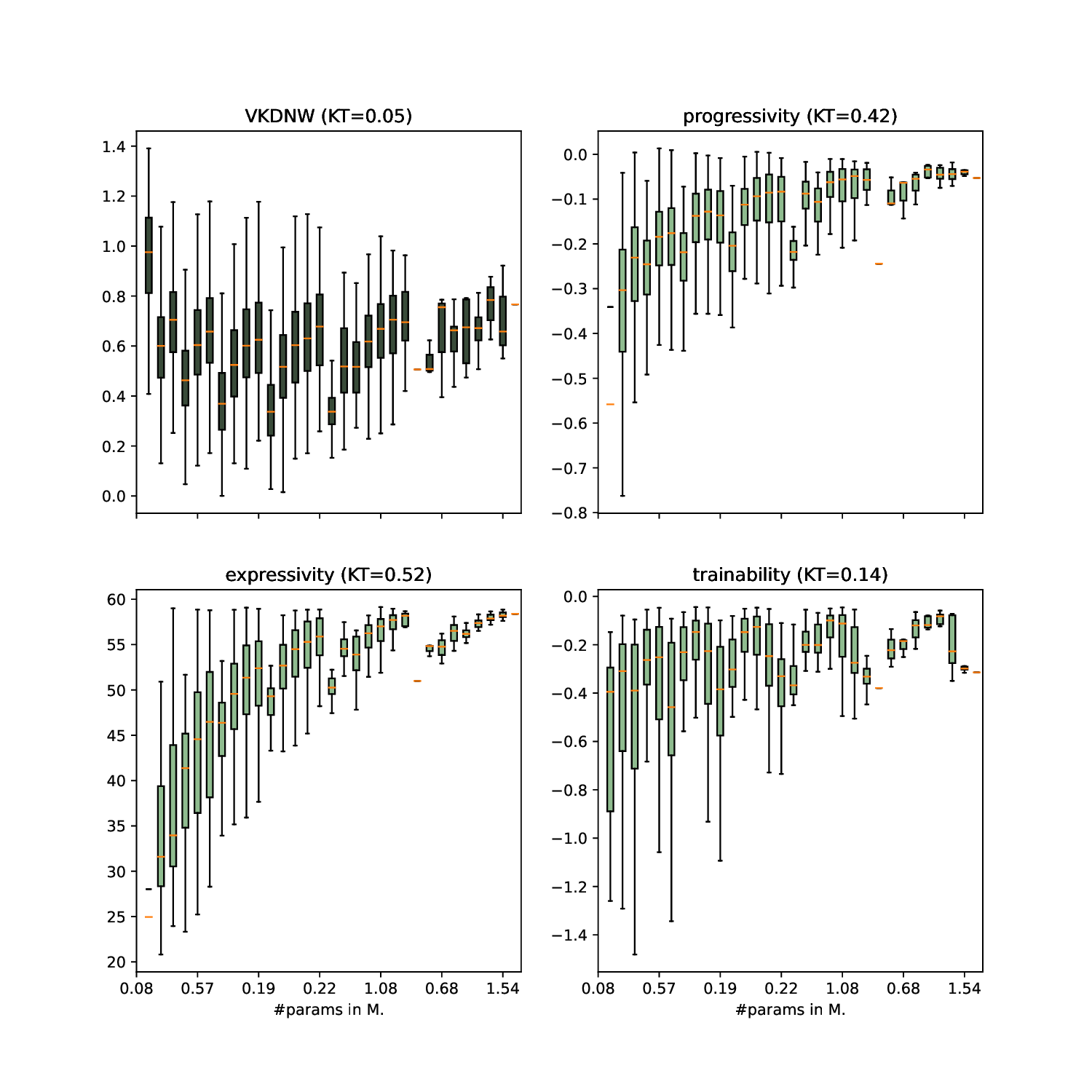}
    \caption{Components of AZ-NAS \cite{lee2024az} and our VKDNW are compared w.r.t. correlation with number of model parameters, in the NAS-Bench-201 search space \cite{dong2020bench} on ImageNet16-120 \cite{chrabaszcz2017downsampled} dataset. Our VKDNW proxy has the lowest correlation, ie. is the most invariant to the size of the model.\vspace{-15pt}}
    \label{fig:boxplot_per_trainable_parameters}
\end{figure}
\paragraph{Orthogonality of VKDNW.}

Our score VKDNW is based on information orthogonal to the size of the network: in \Cref{fig:boxplot_per_trainable_layers}, we show that unlike previous work, VKDNW is not correlated with the network size measured by $\aleph$ (number of trainable layers). In \Cref{fig:boxplot_per_trainable_parameters}, we present similar results where we now measure the network size by the number of trainable parameters. We can see that VKDNW keeps the orthogonality property even after change of the size proxy.

\begin{table}[h]
    \centering

\begin{tabular}{|ccccc|ccc|}
\hline
V & J & E & T & F & (KT) & (SPR) & (nDCG) \\
 \hline
 \checkmark & & & & & \textbf{0.622} & \textbf{0.814} & \textbf{0.608} \\
 & \checkmark & & & & 0.603 & 0.781 & 0.339 \\
 & & \checkmark & & & 0.588 & 0.779 & 0.274 \\
& & & \checkmark & & 0.353 & 0.517 & 0.233 \\
& & & & \checkmark & 0.545 & 0.718 & 0.403 \\
\hline
\checkmark & \checkmark & & & & 0.677 & 0.851 & 0.565 \\
\checkmark & & \checkmark & & & 0.675 & 0.851 & 0.489 \\
\checkmark & & & \checkmark & & 0.622 & 0.821 & 0.552 \\
\checkmark & & & & \checkmark & 0.619 & 0.811 & 0.557 \\
& \checkmark & \checkmark & & & 0.630 & 0.815 & 0.349 \\
& \checkmark & & \checkmark & & 0.621 & 0.818 & 0.505 \\
& \checkmark & & & \checkmark & 0.695 & 0.863 & 0.574 \\
& & \checkmark & \checkmark & & 0.616 & 0.818 & 0.463 \\
& & \checkmark & & \checkmark & 0.642 & 0.818 & 0.434 \\
& & & \checkmark & \checkmark & 0.617 & 0.815 & 0.580 \\
\hline
\checkmark & \checkmark & & \checkmark & & 0.698 & 0.879 & 0.632 \\
\checkmark & \checkmark & \checkmark & & & 0.686 & 0.858 & 0.515 \\
\checkmark & \checkmark & & & \checkmark & 0.696 & 0.868 & 0.616 \\
\checkmark & & \checkmark & \checkmark & & 0.698 & 0.882 & 0.612 \\
\checkmark & & \checkmark & & \checkmark & 0.672 & 0.848 & 0.512 \\
\checkmark & & & \checkmark & \checkmark & 0.681 & 0.870 & 0.675 \\
& \checkmark & \checkmark & \checkmark & & 0.674 & 0.862 & 0.527 \\
& \checkmark & \checkmark & & \checkmark & 0.695 & 0.859 & 0.484 \\
& \checkmark & & \checkmark & \checkmark & 0.726 & 0.899 & 0.673 \\
& & \checkmark & \checkmark & \checkmark & 0.702 & 0.883 & 0.630 \\
\hline
\checkmark & \checkmark & \checkmark & \checkmark & & 0.717 & 0.891 & 0.623 \\
\checkmark & \checkmark & \checkmark & & \checkmark & 0.706 & 0.871 & 0.553 \\
\checkmark & \checkmark & & \checkmark & \checkmark & 0.736 & 0.905 & 0.695 \\
\checkmark & & \checkmark & \checkmark & \checkmark & 0.722 & 0.896 & 0.658 \\
& \checkmark & \checkmark & \checkmark & \checkmark & 0.735 & 0.901 & 0.646 \\
\hline
\checkmark & \checkmark & \checkmark & \checkmark & \checkmark & \textbf{0.743} & \textbf{0.906} & \textbf{0.664} \\
\hline
\end{tabular}
\caption{Components of $\text{VKDNW}_{\text{agg}}$ rank with non-linear aggregation. Consistency is shown with respect to Kendall's $\tau$ (KT), Spearman's $\rho$ (SPR) and Normalized Discounted Cumulative Gain ($\text{nDCG}$) with $P=1000$ on ImageNet16-120 image dataset \cite{chrabaszcz2017downsampled}. Here V, J, E, T and F stand for $\text{VKDNW}_{\text{single}}$, Jacov, expressivity, trainability and FLOPs respectively (see Sec. \ref{subsec:ranking}).
}
\label{tab:ablationComponentsFull}
\end{table}
\paragraph{Components of the aggregated rank.}
Our aggregated rank $\text{VKDNW}_{\text{agg}}$ combines information from 5 different sources: our $\text{VKDNW}_{\text{single}}$, Jacov, expressivity, trainability and FLOPs (see \cref{subsec:ranking}). In \Cref{tab:ablationComponentsFull}, all $2^5$ combinations of keeping/dropping every of the 5 sources are evaluated on ImageNet16-120. We can see that our ranking $\text{VKDNW}_{\text{single}}$ is the strongest component as it has the highest marginal performance in all three considered metrics. The lowest performance drop is observed for expressivity: without this component the method would even perform better in the (nDCG) metric than the original variant $\text{VKDNW}_{\text{agg}}$. We decided to include expressivity in the final ranking as we optimized for all three metrics (KT), (SPR) and (nDCG) simultaneously.

\end{document}